\documentclass[11pt,a4paper]{article}
\usepackage[utf8]{inputenc}
\usepackage[hyperref]{acl2020}
\usepackage[arabic, english]{babel}

\usepackage[LFE,LAE]{fontenc}
\usepackage{times}
\usepackage{latexsym}
\usepackage[OT2,T1]{fontenc}
\usepackage{tipa}
\usepackage{arabtex}
\usepackage{url}
\usepackage{amssymb}
\usepackage{amsmath}

\usepackage{comment}

\usepackage{multirow}
\usepackage{caption}
\usepackage{rotating}
\usepackage{graphicx}
\usepackage{color}
\usepackage{placeins}
\usepackage{bbm}
\usepackage{sidecap}
\usepackage{wrapfig}
\usepackage{booktabs}

\newcommand\textcyr[1]{{\fontencoding{OT2}\fontfamily{wncyr}\selectfont #1}}

\newcommand\russianchar[1]{\textcyr{\textit{#1}}}
\newcommand\russianword[1]{\textcyr{#1}}
\newcommand\sci[1]{{\small[{#1}]}}

\newcommand{\Sref}[1]{\S\ref{#1}}

\newcommand{\Fref}[1]{Figure~\ref{#1}}

\newcommand{\Tref}[1]{Table~\ref{#1}}

\aclfinalcopy 
\def\aclpaperid{3141} 

 \usepackage[disable]{todonotes}

\title{Phonetic and Visual Priors for Decipherment of Informal Romanization}

\author{Maria Ryskina$^1$\quad Matthew R. Gormley$^2$\quad Taylor Berg-Kirkpatrick$^3$ \\
         $^1$Language Technologies Institute, Carnegie Mellon University \\ 
         $^2$Machine Learning Department, Carnegie Mellon University \\ 
         $^3$Computer Science and Engineering, University of California, San Diego \\
         \texttt{\{mryskina,mgormley\}@cs.cmu.edu} \quad \texttt{tberg@eng.ucsd.edu}}

\date{}

\begin{document}
\maketitle
\begin{abstract}
  \emph{Informal romanization} is an idiosyncratic process used by humans in informal digital communication to encode non-Latin script languages into Latin character sets found on common keyboards. Character substitution choices differ between users but have been shown to be governed by the same main principles observed across a variety of languages---namely, character pairs are often associated through phonetic or visual similarity. We propose a noisy-channel WFST cascade model for deciphering the original non-Latin script from observed romanized text in an unsupervised fashion. We train our model directly on romanized data from two languages: Egyptian Arabic and Russian. We demonstrate that adding inductive bias through phonetic and visual priors on character mappings substantially improves the model's performance on both languages, yielding results much closer to the supervised skyline. Finally, we introduce a new dataset of romanized Russian, collected from a Russian social network website and partially annotated for our experiments.\footnote{The code and data are available at \url{https://github.com/ryskina/romanization-decipherment}}
\end{list}
\end{abstract}

\section{Introduction}

Written online communication poses a number of challenges for natural language processing systems, including the presence of neologisms, code-switching, and the use of non-standard orthography. One notable example of orthographic variation in social media is \emph{informal romanization}\footnote{Our focus on \emph{informal} transliteration excludes formal settings such as pinyin for Mandarin where transliteration conventions are well established.}---speakers of languages written in non-Latin alphabets encoding their messages in Latin characters, for convenience or due to technical constraints (improper rendering of native script or keyboard layout incompatibility). An example of such a sentence can be found in \Fref{fig:example}. Unlike named entity transliteration where the change of script represents the change of language, here Latin characters serve as an intermediate symbolic representation to be decoded by another speaker of the same source language, calling for a completely different transliteration mechanism: instead of expressing the pronunciation of the word according to the phonetic rules of another language, informal transliteration can be viewed as a substitution cipher, where each source character is replaced with a similar Latin character.

In this paper, we focus on decoding informally romanized texts back into their original scripts. We view the task as a decipherment problem and propose an unsupervised approach, which allows us to save annotation effort since parallel data for informal transliteration does not occur naturally. We propose a weighted finite-state transducer (WFST) cascade model that learns to decode informal romanization without parallel text, relying only on transliterated data and a language model over the original orthography. We test it on two languages, Egyptian Arabic and Russian, collecting our own dataset of romanized Russian from a Russian social network website \texttt{vk.com.}

\begin{figure}[]
\centering
\includegraphics[width=\columnwidth]{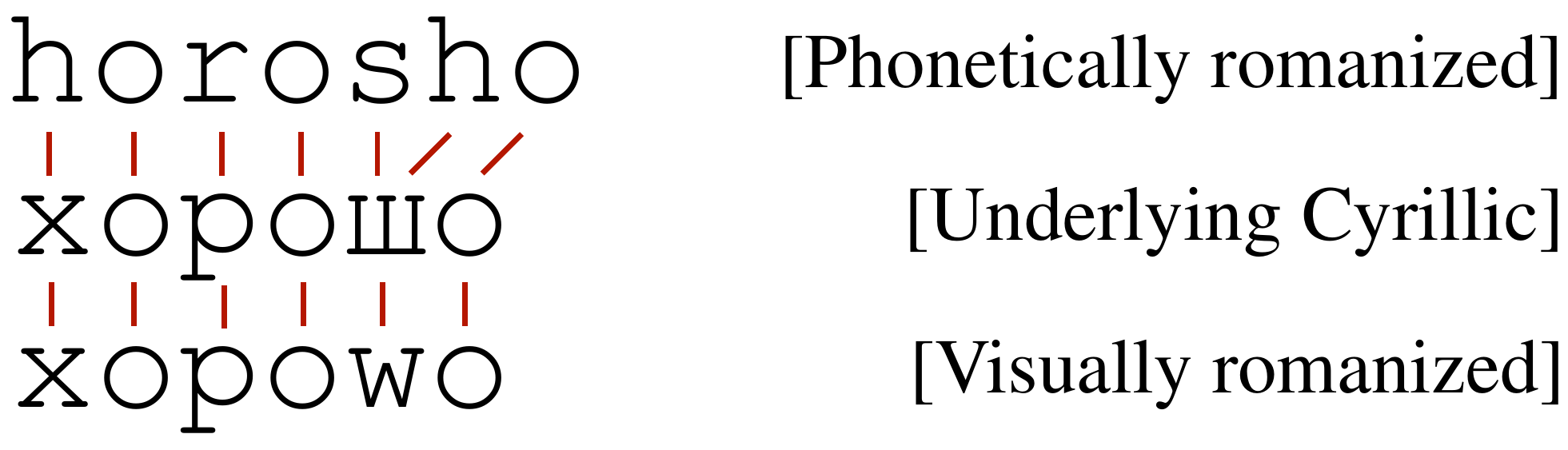}
  \caption{\label{fig:horosho} Example transliterations of a Russian word \russianword{horosho} \sci{horo\v so, `good'} (middle) based on phonetic (top) and visual (bottom) similarity, with character alignments displayed. The phonetic-visual dichotomy gives rise to one-to-many mappings such as \russianchar{sh}~\textipa{/S/}~$\rightarrow$~\emph{sh~/~w}.}
\end{figure}

\begin{figure}[t]
\noindent {4to mowet bit' ly4we?} \hfill [Romanized]\\
\russianword{Chto mozhet bytp1 luchxe?}  \hfill [Latent Cyrillic]\\
{\v Cto mo\v zet byt' lu\v c\v se?} \hfill [Scientific]\\
\textipa{/Sto "moZ\textbari t b\textbari t\textsuperscript{j} "lu\texttoptiebar{tS}S\textbari/} \hfill [IPA]\\
{What can be better?} \hfill [Translated]
\caption{Example of an informally romanized sentence from the dataset presented in this paper, containing a many-to-one mapping \russianchar{zh}~\emph{/}~\russianchar{sh}~$\rightarrow$~\emph{w}. Scientific transliteration, broad phonetic transcription, and translation are not included in the dataset and are presented for illustration only.\label{fig:example}}
\end{figure}

Since informal transliteration is not standardized, converting romanized text back to its original orthography requires reasoning about the specific user's transliteration preferences and handling many-to-one (\Fref{fig:example}) and one-to-many (\Fref{fig:horosho}) character mappings, which is beyond traditional rule-based converters. Although user behaviors vary, there are two dominant patterns in informal romanization that have been observed independently across different languages, such as Russian~\cite{paulsen20149}, dialectal Arabic~\cite{darwish2014arabizi} or Greek~\cite{chalamandaris2006all}:
\\[0.05in]
\noindent {\bf Phonetic similarity:} Users represent source characters with Latin characters or digraphs associated with similar phonemes (e.g.\ \russianchar{m}~\textipa{/m/}~$\rightarrow$~\emph{m}, \russianchar{l}~\textipa{/l/}~$\rightarrow$~\emph{l} 
in \Fref{fig:example}). This substitution method requires implicitly tying the Latin characters to a phonetic system of an intermediate language (typically, English).
\\[0.05in]
\noindent {\bf Visual similarity:} Users replace source characters with similar-looking symbols (e.g.\
\russianchar{ch}~\textipa{/\texttoptiebar{tS\textsuperscript{j}}/}~$\rightarrow$~\emph{4}, 
\russianchar{u}~\textipa{/u/}~$\rightarrow$~\emph{y} 
in \Fref{fig:example}). Visual similarity choices often involve numerals, especially when the corresponding source language phoneme has no English equivalent (e.g.\ Arabic \AR{ع}~\textipa{/Q/}~$\rightarrow$~3).

Taking that consistency across languages into account, we show that incorporating these style patterns into our model as priors on the emission parameters---also constructed from naturally occurring resources---improves the decoding accuracy on both languages. We compare the proposed unsupervised WFST model with a supervised WFST, an unsupervised neural architecture, and commercial systems for decoding romanized Russian (\emph{translit}) and Arabic (\emph{Arabizi}). Our unsupervised WFST outperforms the unsupervised neural baseline on both languages.

\section{Related work}

Prior work on informal transliteration uses supervised approaches with character substitution rules either manually defined or learned from automatically extracted character alignments~\cite{darwish2014arabizi, chalamandaris2004bypassing}. Typically, such approaches are pipelined: they produce candidate transliterations and rerank them using modules encoding knowledge of the source language, such as morphological analyzers or word-level language models~\cite{al2014automatic, eskander2014foreign}. Supervised finite-state approaches have also been explored \citep{wolf2019latin, hellsten2017transliterated}; these WFST cascade models are similar to the one we propose, but they encode a different set of assumptions about the transliteration process due to being designed for abugida scripts (using consonant-vowel syllables as units) rather than alphabets. To our knowledge, there is no prior unsupervised work on this problem.

Named entity transliteration, a task closely related to ours, is better explored, 
but there is little unsupervised work on this task as well. In particular, \citet{ravi2009learning} propose a fully unsupervised version of the WFST approach introduced by \citet{knight1998machine}, reframing the task as a decipherment problem and learning cross-lingual phoneme mappings from monolingual data. We take a similar path, although it should be noted that named entity transliteration methods cannot be straightforwardly adapted to our task due to the different nature of the transliteration choices. The goal of the standard transliteration task is to communicate the pronunciation of a sequence in the source language (SL) to a speaker of the target language (TL) by rendering it appropriately in the TL alphabet; in contrast, informal romanization emerges in communication between SL speakers only, and TL is not specified. If we picked any specific Latin-script language to represent TL (e.g.\ English, which is often used to ground phonetic substitutions), many of the informally romanized sequences would still not conform to its pronunciation rules: the transliteration process is character-level rather than phoneme-level and does not take possible TL digraphs into account (e.g.\ Russian \russianchar{skh}~\textipa{/sx/}~$\rightarrow$~\emph{sh}), and it often involves eclectic visual substitution choices such as numerals or punctuation (e.g.\ Arabic  \AR{تحت}~\sci{tHt,~`under'}\footnote{The square brackets following a foreign word show its linguistic transliteration (using the scientific and the Buckwalter schemas for Russian and Arabic respectively) and its English translation.} $\rightarrow$ \emph{ta7t}, Russian \russianword{dlya}~\sci{dlja,~`for'} $\rightarrow$ \emph{dl9|} ).

Finally, another relevant task is translating between closely related languages, possibly written in different scripts. An approach similar to ours is proposed by \citet{pourdamghani2017deciphering}. They also take an unsupervised decipherment approach: the cipher model, parameterized as a WFST, is trained to encode the source language character sequences into the target language alphabet as part of a character-level noisy-channel model, and at decoding time it is composed with a word-level language model of the source language. 
Recently, the unsupervised neural architectures~\cite{lample2018unsupervised, lample2018multipleattribute} have also been used for related language translation and similar decipherment tasks~\cite{he2020a}, and we extend one of these neural models to our character-level setup to serve as a baseline (\Sref{sec:experiments}).

\section{Methods}

We train a character-based noisy-channel model that transforms a character sequence $o$ in the native alphabet of the language into a sequence of Latin characters $l$, and use it to decode the romanized sequence $l$ back into the original orthography. Our proposed model is composed of separate transition and emission components as discussed in \Sref{sec:model}, similarly to an HMM. However, an HMM assumes a one-to-one alignment between the characters of the observed and the latent sequences, which is not true for our task. One original script character can be aligned to two consecutive Latin characters or vice versa: for example, when a phoneme is represented with a single symbol on one side but with a digraph on the other (\Fref{fig:horosho}), or when a character is omitted on one side but explicitly written on the other (e.g.\ short vowels not written in unvocalized Arabic but written in transliteration, or the Russian soft sign \russianchar{p1} representing palatalization being often omitted in the romanized version). To handle those alignments, we introduce insertions and deletions into the emission model and modify the emission transducer to limit the number of consecutive insertions and deletions. In our experiments, we compare the performance of the model with and without informative phonetic and visual similarity priors described in \Sref{sec:prior}.

\subsection{Model \label{sec:model}}

If we view the process of romanization as encoding a source sequence $o$ into Latin characters, we can consider each observation $l$ to have originated via $o$ being generated from a distribution $p(o)$ and then transformed to Latin script according to another distribution $p(l|o)$. We can write the probability of the observed Latin sequence as:
\begin{equation}
    p(l) = \sum_{o} p(o; \gamma) \cdot p(l|o; \theta) \cdot p_{\text{prior}}(\theta; \alpha)
    \label{eq:noisychannel}
\end{equation}
The first two terms in~\eqref{eq:noisychannel} correspond to the probabilities under the transition model (the language model trained on the original orthography) and the emission model respectively. The third term represents the prior distribution on the emission model parameters through which we introduce human knowledge into the model. Our goal is to learn the parameters $\theta$ of the emission distribution with the transition parameters $\gamma$ being fixed.

We parameterize the emission and transition distributions as weighted finite-state transducers (WFSTs):
\\[0.05in]
\noindent\textbf{Transition WFSA} The n-gram weighted finite-state acceptor (WFSA) $T$ represents a character-level n-gram language model of the language in the native script, producing the native alphabet character sequence $o$ with the probability $p(o; \gamma)$. We use the parameterization of~\citet{allauzen2003generalized}, with the states encoding conditioning history, arcs weighted by n-gram probabilities, and failure transitions representing backoffs. The role of $T$ is to inform the model of what well-formed text in the original orthography looks like; its parameters $\gamma$ are learned from a separate corpus and kept fixed during the rest of the training.
\\[0.05in]
\noindent\textbf{Emission WFST} The emission WFST $S$ transduces the original script sequence $o$ to a Latin sequence $l$ with the probability $p(l|o; \theta)$. Since there can be multiple paths through $S$ that correspond to the input-output pair $(o, l)$, this probability is summed over all such paths (i.e. is a marginal over all possible monotonic character alignments):
\begin{align}
    p(l|o; \theta) = \sum_e p(l, e | o; \theta)
\end{align}
We view each path $e$ as a sequence of edit operations: substitutions of original characters with Latin ones ($c_o \rightarrow c_l$), insertions of Latin characters ($\epsilon \rightarrow c_l$), and deletions of original characters ($c_o \rightarrow \epsilon$). Each arc in $S$ corresponds to one of the possible edit operations; an arc representing the edit $c_o \rightarrow c_l$ is characterized by the input label $c_o$, the output label $c_l$, and the weight $-\log p(c_l | c_o; \theta)$. The emission parameters $\theta$ are the multinomial conditional probabilities of the edit operations $p(c_l | c_o)$; we learn $\theta$ using the algorithm described in \Sref{sec:learn}.

\begin{SCfigure*}
\centering
\includegraphics[scale=0.5]{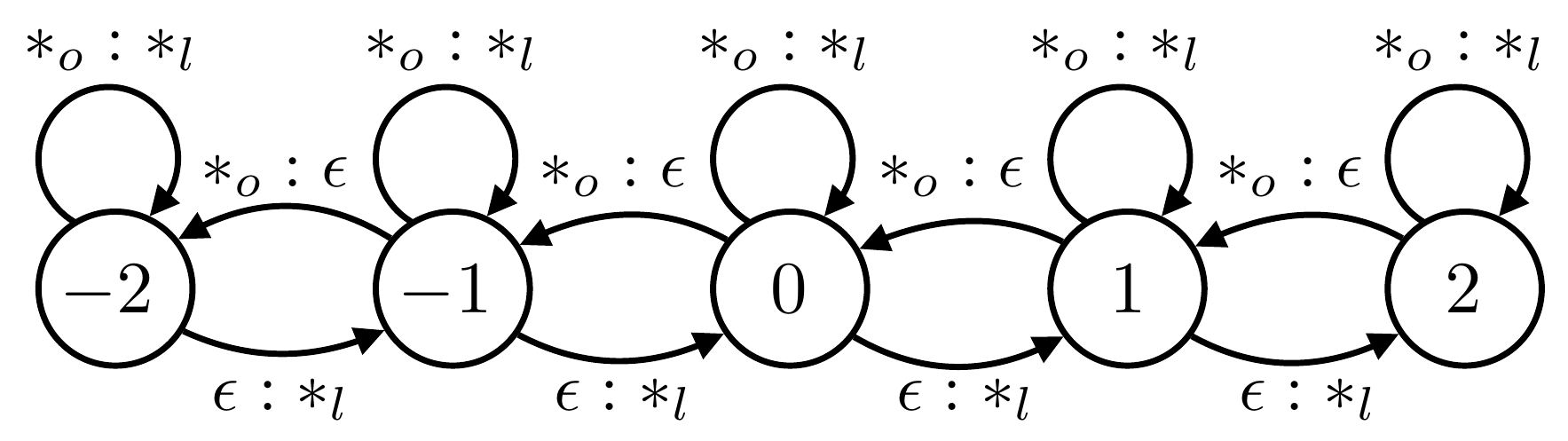}
  \caption{Schematic of the emission WFST with limited delay (here, up to 2) with states labeled by their delay values. $*_o$ and $*_l$ represent an arbitrary original or Latin symbol respectively.
  Weights of the arcs are omitted for clarity; weights with the same input-output label pairs are tied. \label{fig:fst}}
\end{SCfigure*}

\subsection{Phonetic and visual priors \label{sec:prior}}

To inform the model of which pairs of symbols are close in the phonetic or visual space, we introduce the priors on the emission parameters, increasing the probability of an original alphabet character being substituted by a similar Latin one. Rather than attempting to operationalize the notions of phonetic or visual similarity, we choose to read the likely mappings between symbols off human-compiled resources that use the same underlying principle: phonetic keyboard layouts and visually confusable symbol lists. Examples of mappings that we encode as priors can be found in~\Tref{tab:prior}.
\\[0.1in]
\noindent \textbf{Phonetic similarity}\quad Since we think of the informal romanization as a cipher, we aim to capture the phonetic similarity between characters based on association rather than on the actual grapheme-to-phoneme mappings in specific words. We approximate it using \emph{phonetic keyboard layouts}, one-to-one mappings built to bring together ``similar-sounding'' characters in different alphabets. We take the character pairs from a union of multiple layouts for each language, two for Arabic\footnote{\url{http://arabic.omaralzabir.com/}, \url{https://thomasplagwitz.com/2013/01/06/imrans-phonetic-keyboard-for-arabic/}} and four for Russian.\footnote{\url{http://winrus.com/kbd_e.htm}}
The main drawback of using keyboard layouts is that they require every character to have a Latin counterpart, so some mappings will inevitably be arbitrary; we compensate for this effect by averaging over several layouts.
\\[0.1in]
\noindent \textbf{Visual similarity}\quad The strongest example of visual character similarity would be \emph{homoglyphs}---symbols from different alphabets represented by the same glyph, such as Cyrillic~\russianchar{a} and Latin~\textit{a}. The fact that homoglyph pairs can be made indistinguishable in certain fonts has been exploited in phishing attacks, e.g.\ when Latin characters are replaced by virtually identical Cyrillic ones \cite{gabrilovich2002homograph}. This led the Unicode Consortium to publish a list of symbols and symbol combinations similar enough to be potentially confusing to the human eye (referred to as \emph{confusables}).\footnote{\url{https://www.unicode.org/Public/security/latest/confusables.txt}} This list contains not only exact homoglyphs but also strongly homoglyphic pairs such as Cyrillic \russianchar{Yu} and Latin \textit{lO}. 

We construct a visual prior for the Russian model from all Cyrillic--Latin symbol pairs in the Unicode confusables list.\footnote{In our parameterization, we cannot introduce a mapping from one to multiple symbols or vice versa, so we map all possible pairs instead:  (\russianchar{yu}, \textit{lo}) $\rightarrow$ (\russianchar{yu}, \textit{l}), (\russianchar{yu}, \textit{o}).} Although this list does not cover more complex visual associations used in informal romanization, such as partial similarity (Arabic Alif with Hamza \AR{أ}~$\rightarrow$~2 due to Hamza \AR{ء}\enspace resembling an inverted 2) or similarity conditioned on a transformation such as reflection (Russian \russianchar{l}~$\rightarrow$~\textit{v}), it makes a sensible starting point. However, this restrictive definition of visual similarity does not allow us to create a visual prior for Arabic---the two scripts are dissimilar enough that the confusables list does not contain any Arabic--Latin character pairs. Proposing a more nuanced definition of visual similarity for Arabic and the associated prior is left for future work.\\

We incorporate these mappings into the model as Dirichlet priors on the emission parameters: $\theta \sim \text{Dir}(\alpha)$, where each dimension of the parameter $\alpha$ corresponds to a character pair $(c_o, c_l)$, and the corresponding element of $\alpha$ is set to the number of times these symbols are mapped to each other in the predefined mapping set.

\begin{table}[t]
    \centering
    \begin{tabular}{lll}
    \toprule
    \multirow{2}{*}{Original} & \multicolumn{2}{c}{Latin} \\
     & Phon. & Vis. \\
     \midrule
     \russianchar{r} \textipa{/r/}  & \emph{r} & \emph{p}  \\
     \russianchar{b} \textipa{/b/} & \emph{b} & \emph{b, 6} \\
     \russianchar{v} \textipa{/v/} & \emph{v, w} & \emph{b} \\
     \midrule
     \AR{و} \textipa{/w, u:, o:/}  & \emph{w, u} & ---  \\
     \AR{خ}  \textipa{/x/} & \emph{k, x} & --- \\
    \bottomrule
    \end{tabular}
    \caption{Example Cyrillic--Latin and Arabic--Latin mappings encoded in the visual and phonetic priors respectively.}
    \label{tab:prior}
\end{table}

\subsection{Learning \label{sec:learn}}

We learn the emission WFST parameters in an unsupervised fashion, observing only the Latin side of the training instances. The marginal likelihood of a romanized sequence $l$ can be computed by summing over the weights of all paths through a lattice obtained by composing $T \circ S \circ A(l)$. Here $A(l)$ is an unweighted acceptor of $l$, which, when composed with a lattice, constrains all paths through the lattice to produce $l$ as the output sequence. The expectation--maximization (EM) algorithm is commonly used to maximize marginal likelihood; however, the size of the lattice would make the computation prohibitively slow. We combine online learning~\cite{liang2009online} and curriculum learning~\cite{bengio2009curriculum} to achieve faster convergence, as described in \Sref{sec:uns}.

\subsubsection{Unsupervised learning \label{sec:uns}}

We use a version of the stepwise EM algorithm described by~\citet{liang2009online}, reminiscent of the stochastic gradient descent in the space of the sufficient statistics. Training data is split into mini-batches, and after processing each mini-batch we update the overall vector of the sufficient statistics $\mu$ and re-estimate the parameters based on the updated vector. The update is performed by interpolating between the current value of the overall vector and the vector of sufficient statistics $s_k$ collected from the $k$-th mini-batch: $\mu^{(k+1)} \leftarrow (1 - \eta_k) \mu^{(k)} + \eta_k s_k$. The stepsize is gradually decreased, causing the model to make smaller changes to the parameters as the learning stabilizes. Following~\citet{liang2009online}, we set it to $\eta_k = (k+2)^{-\beta}$.

However, if the mini-batch contains long sequences, summing over all paths in the corresponding lattices could still take a long time. As we know, the character substitutions are not arbitrary: each original alphabet symbols is likely to be mapped to only a few Latin characters, which means that most of the paths through the lattice would have very low probabilities. We prune the improbable arcs in the emission WFST while training on batches of shorter sentences. Doing this eliminates up to 66\% and up to 76\% of the emission arcs for Arabic and Russian respectively.

We discourage excessive use of insertions and deletions by keeping the corresponding probabilities low at the early stages of training: during the first several updates, we freeze the deletion probabilities at a small initial value and disable insertions completely to keep the model locally normalized. We also iteratively increase the language model order as learning progresses. Once most of the emission WFST arcs have been pruned, we can afford to compose it with a larger language model WFST without the size of the resulting lattice rendering the computation impractical. The two steps of the EM algorithm are performed as follows:
\\[0.05in]
\noindent\textbf{E-step}\quad At the E-step we compute the sufficient statistics for updating $\theta$, which in our case would be the expected number of traversals of each of the emission WFST arcs. For ease of bookkeeping, we compute those expectations using finite-state methods in the expectation semiring~\cite{eisner2002parameter}. Summing over all paths in the lattice is usually performed via shortest distance computation in log semiring; in the expectation semiring, we augment the weight of each arc with a basis vector, where the only non-zero element corresponds to the index of the emission edit operation associated with the arc (i.e. the input-output label pair). This way the shortest distance algorithm yields not only the marginal likelihood but also the vector of the sufficient statistics for the input sequence. 

To speed up the shortest distance computation, we shrink the lattice by limiting delay of all paths through the emission WFST. Delay of a path is defined as the difference between the number of the epsilon labels on the input and output sides of the path. \Fref{fig:fst} shows the schema of the emission WFST where delay is limited. Substitutions are performed without a state change, and each deletion or insertion arc transitions to the next or previous state respectively. When the first (last) state is reached, further deletions (insertions) are no longer allowed.
\\[0.05in]
\noindent\textbf{M-step}\quad The M-step then corresponds to simply re-estimating $\theta$ by appropriately normalizing the obtained expected counts.

\subsubsection{Supervised learning \label{sec:sup}}

We also compare the performance of our model with the same model trained in a supervised way, using the annotated portion of the data that contains parallel $o$ and $l$ sequences. In the supervised case we can additionally constrain the lattice with an acceptor of the original orthography sequence: $A(o) \circ T \circ S \circ A(l)$. However, the alignment between the symbols in $o$ and $l$ is still latent. To optimize this marginal likelihood we still employ the EM algorithm. As this constrained lattice is much smaller, we can run the standard EM without the modifications discussed in \Sref{sec:uns}.

\subsection{Decoding} 

Inference at test time is also performed using finite-state methods and closely resembles the E-step of the unsupervised learning: given a Latin sequence $l$, we construct the machine $T \circ S \circ A(l)$ in the tropical semiring and run the shortest path algorithm to obtain the most probable path $\hat{e}$; the source sequence $\hat{o}$ is read off the obtained path.

\section{Datasets}

Here we discuss the data used to train the unsupervised model. Unlike Arabizi, which has been explored in prior work due to its popularity in the modern online community, a dataset of informally romanized Russian was not available, so we collect and partially annotate our own dataset from the Russian social network \texttt{vk.com}.

\begin{table}[t]
    \centering
    \begin{tabular}{@{}lrrrr@{}}
        \toprule
        & \multicolumn{2}{c}{Arabic} & \multicolumn{2}{c}{Russian}  \\
        & \multicolumn{1}{c}{Sent.} & \multicolumn{1}{c}{Char.} & \multicolumn{1}{c}{Sent.} & \multicolumn{1}{c}{Char.}\\
        \midrule
       LM train & 49K & 935K & 307K & 111M\\
       Train  & 5K & 104K & 5K & 319K\\
       Validation  & 301 & 8K & 227 & 15K\\
       Test & 1K & 20K & 1K & 72K \\
    \bottomrule
    \end{tabular}
    \caption{Splits of the Arabic and Russian data used in our experiments. All Arabic data comes from the LDC BOLT Phase 2 corpus, in which all sentences are annotated with their transliteration into the Arabic script. For the experiments on Russian, the language model is trained on a section of the Taiga corpus, and the train, validation, and test portions are collected by the authors; only the validation and test sentences are annotated.}
    \label{tab:splits}
\end{table}

\subsection{Arabic}

We use the Arabizi portion of the LDC BOLT Phase 2 SMS/Chat dataset~\cite{bies2014transliteration, song2014collecting}, a collection of written informal conversations in romanized Egyptian Arabic annotated with their Arabic script representation. To prevent the annotators from introducing orthographic variation inherent to dialectal Arabic, compliance with the Conventional orthography for dialectal Arabic \citep[CODA;][]{habash2012conventional} is ensured. However, the effects of some of the normalization choices (e.g.\ expanding frequent abbreviations) would pose difficulties to our model.
 
To obtain a subset of the data better suited for our task, we discard any instances which are not originally romanized (5\% of all data), ones where the Arabic annotation contains Latin characters (4\%), or where emoji/emoticon normalization was performed (12\%). The information about the splits is provided in \Tref{tab:splits}. Most of the data is allocated to the language model training set in order to give the unsupervised model enough signal from the native script side. We choose to train the transition model on the annotations from the same corpus to make the language model specific to both the informal domain and the CODA orthography.

\subsection{Russian}

We collect our own dataset of romanized Russian text from a social network website \texttt{vk.com}, adopting an approach similar to the one described by~\citet{darwish2014arabizi}. We take a list of the 50 most frequent Russian lemmas~\cite{lyashevskaya2009chastotnyj}, filtering out those shorter than 3 characters, and produce a set of candidate romanizations for each of them to use as queries to the \texttt{vk.com} API. In order to encourage diversity of romanization styles in our dataset, we generate the queries by defining all plausible visual and phonetic mappings for each Cyrillic character and applying all possible combinations of those substitutions to the underlying Russian word. We scrape public posts on the user and group pages, retaining only the information about which posts were authored by the same user, and manually go over the collected set to filter out coincidental results.

Our dataset consists of 1796 wall posts from 1681 users and communities. Since the posts are quite long on average (248 characters, longest ones up to 15K), we split them into sentences using the NLTK sentence tokenizer, with manual correction when needed. The obtained sentences are used as data points, split into training, validation and test according to the numbers in~\Tref{tab:splits}. The average length of an obtained sentence is 65 characters, which is 3 times longer than an average Arabizi sentence; we believe this is due to the different nature of the data (social media posts vs. SMS). Sentences collected from the same user are distributed across different splits so that we observe a diverse set of romanization preferences in both training and testing. Each sentence in the validation and test sets is annotated by one of the two native speaker annotators, following guidelines similar to those designed for the Arabizi BOLT data \cite{bies2014transliteration}. For more details on the annotation guidelines and inter-annotator agreement, see Appendix~\ref{app:data}.

Since we do not have enough annotations to train the Russian language model on the same corpus, we use a separate in-domain dataset. We take a portion of the Taiga dataset~\citep{shavrina2017methodology}, containing 307K comments scraped from the same social network \texttt{vk.com}, and apply the same preprocessing steps as we did in the collection process.

\section{Experiments \label{sec:experiments}}

Here we discuss the experimental setup used to determine how much information relevant for our task is contained in the character similarity mappings, and how it compares to the amount of information encoded in the human annotations. We compare them by evaluating the effect of the informative priors (described in \Sref{sec:prior}) on the performance of the unsupervised model and comparing it to the performance of the supervised model.
\\[0.1in]
\noindent\textbf{Methods}\quad We compare the performance of our model trained in three different setups: unsupervised with a uniform prior on the emission parameters, unsupervised with informative phonetic and visual priors (\Sref{sec:prior}), and supervised. We additionally compare them to a commercial online decoding system for each language (directly encoding human knowledge about the transliteration process) and a character-level unsupervised neural machine translation architecture (encoding no assumptions about the underlying process at all).

We train the unsupervised models with the stepwise EM algorithm as described in \Sref{sec:uns}, performing stochastic updates and making only one pass over the entire training set. The supervised models are trained on the validation set with five iterations of EM with a six-gram transition model. It should be noted that only a subset of the validation data is actually used in the supervised training: if the absolute value of the delay of the emission WFST paths is limited by $n$, we will not be able to compose a lattice for any data points where the input and output sequences differ in length by more than $n$ (those constitute 22\% of the Arabic validation data and 33\% of the Russian validation data for $n=5$ and $n=2$ respectively). Since all of the Arabic data comes annotated, we can perform the same experiment using the full training set; surprisingly, the performance of the supervised model does not improve (see~\Tref{tab:results}).

The online transliteration decoding systems we use are \texttt{translit.net} for Russian and Yamli\footnote{\url{https://www.yamli.com/}} for Arabic. The Russian decoder is rule-based, but the information about what algorithm the Arabic decoder uses is not disclosed.

We take the unsupervised neural machine translation (UNMT) model of~\citet{lample2018unsupervised} as the neural baseline, using the implementation from the codebase of~\citet{he2020a}, with one important difference: since the romanization process is known to be strictly character-level, we tokenize the text into characters rather than words.
\\[0.1in]
\noindent\textbf{Implementation}\quad We use the OpenFst library \citep{openfst} for the implementation of all the finite-state methods, in conjunction with the OpenGrm NGram library~\citep{opengrm} for training the transition model specifically. We train the character-level n-gram models with Witten--Bell smoothing~\citep{witten1991zero} of orders from two to six. Since the WFSTs encoding full higher-order models become very large (for example, the Russian six-gram model has 3M states and 13M arcs), we shrink all the models except for the bigram one using relative entropy pruning~\citep{stolcke1998entropy}. However, since pruning decreases the quality of the language model, we observe most of the improvement in accuracy while training with the unpruned bigram model, and the subsequent order increases lead to relatively minor gains. Hyperparameter settings for training the transition and emission WFSTs are described in Appendix~\ref{app:hyperpar}.

We optimize the delay limit for each language separately, obtaining best results with 2 for Russian and 5 for Arabic. To approximate the monotonic word-level alignment between the original and Latin sequences, we restrict the operations on the space character to only three: insertion, deletion, and substitution with itself. We apply the same to the punctuation marks (with specialized substitutions for certain Arabic symbols, such as ?~$\rightarrow$~\AR{?}). This substantially reduces the number of arcs in the emission WFST, as punctuation marks make up over half of each of the alphabets.
\\[0.1in]
\noindent\textbf{Evaluation} We use character error rate (CER) as our evaluation metric. We compute CER as the ratio of the character-level edit distance between the predicted original script sequence and the human annotation to the length of the annotation sequence in characters.

\section{Results and analysis}

\begin{table}[]
    \centering
    \resizebox{\columnwidth}{!}{
    \begin{tabular}{@{}lll@{}}
         \toprule
         & Arabic & Russian  \\
         \midrule
         Unsupervised: uniform prior  & 0.735 
                                      & 0.660 
                                      \\
         Unsupervised: phonetic prior & 0.377 
                                      & 0.222 
                                      \\
         Unsupervised: visual prior   & \multicolumn{1}{c}{---}  
                                      & 0.372 
                                      \\
         Unsupervised: combined prior & \multicolumn{1}{c}{---} 
                                      & 0.212 
                                      \\
         \midrule
         Supervised                   & 0.225* & 0.140 \\
         UNMT                         & 0.791 & 0.242 \\ 
         Commercial                   & 0.206 & 0.137 \\
         \bottomrule
    \end{tabular}
    }
    \caption{Character error rate for different experimental setups. We compare unsupervised models with and without informative priors with the supervised model (trained on validation data) and a commercial online system. We do not have a visual prior for Arabic due to the Arabic--Latin visual character similarity not being captured by the restrictive confusables list that defines the prior (see \Sref{sec:prior}). Each supervised and unsupervised experiment is performed with 5 random restarts. *The Arabic supervised experiment result is for the model trained on the validation set; training on the 5K training set yields 0.226.}
    \label{tab:results}
\end{table}

The CER values for the models we compare are presented in \Tref{tab:results}. One trend we notice is that the error rate is lower for Russian than for Arabic in all the experiments, including the uniform prior setting, which suggests that decoding Arabizi is an inherently harder task. Some of the errors of the Arabic commercial system could be explained by the decoder predictions being plausible but not matching the CODA orthography of the reference.
\\[0.1in]
\noindent\textbf{Effect of priors}\quad The unsupervised models without an informative prior perform poorly for either language, which means that there is not enough signal in the language model alone under the training constraints we enforce. Possibly, the algorithm could have converged to a better local optimum if we did not use the online algorithm and prune both the language model and the emission model; however, that experiment would be infeasibly slow. Incorporating a phonetic prior reduces the error rate by 0.36 and 0.44 for Arabic and Russian respectively, which provides a substantial improvement while maintaining the efficiency advantage. The visual prior for Russian appears to be slightly less helpful, improving CER by 0.29. We attribute the better performance of the model with the phonetic prior to the sparsity and restrictiveness of the visually confusable symbol mappings, or it could be due to the phonetic substitutions being more popular with users. Finally, combining the two priors for Russian leads to a slight additional improvement in accuracy over the phonetic prior only.

\begin{figure}[t]
    \centering
    \scalebox{0.95}{\includegraphics[width=\columnwidth]{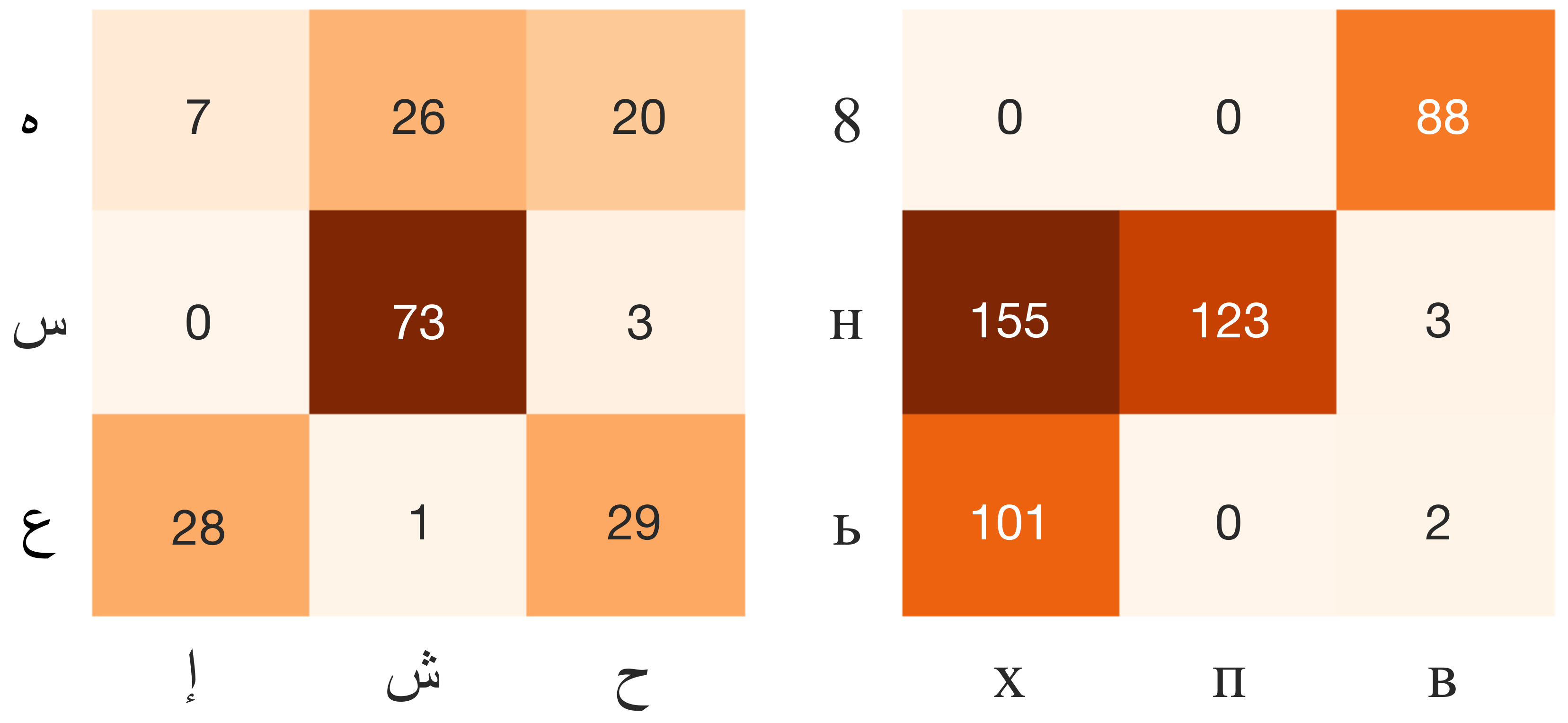}}
    \caption{Fragments of the confusion matrix comparing test time predictions of the best-performing unsupervised models for Arabic (left) and Russian (right) to human annotations. Each number represents the  count of the corresponding substitution in the best alignment (edit distance path) between the predicted and gold sequences, summed over the test set. Rows stand for predictions, columns correspond to ground truth.}
    \label{fig:conf}
\end{figure}

\begin{table}[t]
    \centering
    \begin{tabular}{ll}
    \toprule
    Original & Latin \\
     \midrule
     \russianchar{r} \textipa{/r/}  & \emph{r} (.93), \emph{p} (.05)  \\
     \russianchar{b} \textipa{/b/} & \emph{b} (.95), \emph{6} (.02) \\
     \russianchar{v} \textipa{/v/} & \emph{v} (.87), \emph{8} (.05), \emph{w} (.05) \\
     \midrule
     \AR{و} \textipa{/w, u:, o:/}  & \emph{w} (.48), \emph{o} (.33), \emph{u} (.06)  \\
     \AR{خ}  \textipa{/x/} & \emph{5} (.76), \emph{k} (.24)\\
    \bottomrule
    \end{tabular}
    \caption{Emission probabilities learned by the supervised model (compare to \Tref{tab:prior}). All substitutions with probability greater than 0.01 are shown.\label{tab:params}}
\end{table}
We additionally verify that the phonetic and visual similarity-based substitutions are prominent in informal romanization by inspecting the emission parameters learned by the supervised model with a uniform prior (\Tref{tab:params}). We observe that: (a) the highest-probability substitutions can be explained by either phonetic or visual similarity, and (b) the external mappings we use for our priors are indeed appropriate since the supervised model recovers the same mappings in the annotated data.
\\[0.1in]
\noindent\textbf{Error analysis}\quad \Fref{fig:conf} shows some of the elements of the confusion matrices for the test predictions of the best-performing unsupervised models in both languages. We see that many of the frequent errors are caused by the model failing to disambiguate between two plausible decodings of a Latin character, either mapped to it through different types of similarity (
\russianchar{n}~\textipa{/n/}~[phonetic]~$\rightarrow$~\emph{n}~$\leftarrow$~[visual]~\russianchar{p}, 
\russianchar{n}~[visual]~$\rightarrow$~\emph{h}~$\leftarrow$~[phonetic]~\russianchar{h}~\textipa{/x/}
), or the same one (visual 
8~$\rightarrow$~8~$\leftarrow$~\russianchar{v},
phonetic \AR{ه}~\textipa{/h/}~$\rightarrow$~\emph{h}~$\leftarrow$~\AR{ح}~\textipa{/\textcrh/}
); such cases could be ambiguous for humans to decode as well. 

Other errors in \Fref{fig:conf} illustrate the limitations of our parameterization and the resources we rely on. Our model does not allow one-to-many alignments, which leads to digraph interpretation errors such as \AR{s}~\textipa{/s/}~+~\AR{h}~\textipa{/h/}~$\rightarrow$~\emph{sh}~$\leftarrow$~\AR{ش}~\textipa{/S/}. Some artifacts of the resources our priors are based on also pollute the results: for example, the confusion between  \russianchar{p1} and \russianchar{h} in Russian is explained by the Russian soft sign \russianchar{p1}, which has no English phonetic equivalent, being arbitrarily mapped to the Latin \emph{x} in one of the phonetic keyboard layouts.
\\[0.1in]
\noindent\textbf{Comparison to UNMT}\quad The unsupervised neural model trained on Russian performs only marginally worse than the unsupervised WFST model with an informative prior, demonstrating that with a sufficient amount of data the neural architecture is powerful enough to learn the character substitution rules without the need for the inductive bias. However, we cannot say the same about Arabic---with a smaller training set (see \Tref{tab:splits}), the UNMT model is outperformed by the unsupervised WFST even without an informative prior. The main difference in the performance between the two models comes down to the trade-off between structure and power: although the neural architecture captures long-range dependencies better due to having a stronger language model, it does not provide an easy way of enforcing character-level constraints on the decoding process, which the WFST model encodes by design. As a result, we observe that while the UNMT model can recover whole words more successfully (for Russian it achieves 45.8 BLEU score, while the best-performing unsupervised WFST is at 20.4), it also tends to arbitrarily insert or repeat words in the output, which leads to higher CER.

\section{Conclusion}

This paper tackles the problem of decoding non-standardized informal romanization used in social media into the original orthography without parallel text. We train a WFST noisy-channel model to decode romanized Egyptian Arabic and Russian to their original scripts with the stepwise EM algorithm combined with curriculum learning and demonstrate that while the unsupervised model by itself performs poorly, introducing an informative prior that encodes the notion of phonetic or visual character similarity brings its performance substantially closer to that of the supervised model.

The informative priors used in our experiments are constructed using sets of character mappings compiled for other purposes but using the same underlying principle (phonetic keyboard layouts and the Unicode confusable symbol list). While these mappings provide a convenient way to avoid formalizing the complex notions of the phonetic and visual similarity, they are restrictive and do not capture all the diverse aspects of similarity that idiosyncratic romanization uses, so designing more suitable priors via operationalizing the concept of character similarity could be a promising direction for future work. Another research avenue that could be explored is modeling specific user preferences: since each user likely favors a certain set of character substitutions, allowing user-specific parameters could improve decoding and be useful for authorship attribution.

\section*{Acknowledgments}
This project is funded in part by the NSF under
grants 1618044 and 1936155, and by the NEH under grant HAA256044-17. The authors thank John Wieting, Shruti Rijhwani, David Mortensen, Nikita Srivatsan, and Mahmoud Al Ismail for helpful discussion, Junxian He for help with the UNMT experiments, Stas Kashepava for data annotation, and the three anonymous reviewers for their valuable feedback.

\bibliography{translit_refs}
\bibliographystyle{acl_natbib}

\clearpage
\appendix
\section{Data collection and annotation \label{app:data}}

\noindent\textbf{Preprocessing}\quad We generate a set of 270 candidate transliterations of 26 Russian words to use as queries. However, many of the produced combinations are highly unlikely and yield no results, and some happen to share the spelling with words in other languages (most often other Slavic languages that use Latin script, such as Polish). We scrape public posts on user and group pages, retaining only the information about which posts were authored by the same user, and manually go over the collected set to filter out coincidental results. We additionally preprocess the collected data by normalizing punctuation and removing non-ASCII characters and emoji. We also replace all substrings of the same character repeated more than twice to only two repetitions, as suggested by~\citet{darwish2012language}, since these repetitions are more likely to be a written expression of emotion than to be explained by the underlying Russian sentence. The same preprocessing steps are applied to the original script side of the data (the annotations and the monolingual language model training corpus) as well.
\\[0.07in]
\noindent\textbf{Annotation guidelines}\quad While transliterating, annotators perform orthographic normalization wherever possible, correcting typos and errors in word boundaries; grammatical errors are not corrected. Tokens that do not require transliteration (foreign words, emoticons) or ones that annotator fails to identify (proper names, badly misspelled words) are removed from the romanized sentence and not transliterated. Although it means that some of the test set sentences will not exactly represent the original romanized sequence, it will help us ensure that we are only testing our model's ability to transliterate rather than make word-by-word normalization decisions.

In addition, 200 of the validation sequences are dually annotated to measure the inter-annotator agreement. We evaluate it using character error rate (CER; edit distance between the two sequences normalized by the length of the reference sequence), the same metric we use to evaluate the model's performance. In this case, since neither of the annotations is the ground truth, we compute CER in both directions and average. Despite the discrepancies caused by the annotators deleting unknown words at their discretion, average CER is only 0.014, which indicates a very high level of agreement.

\section{Hyperparameter settings \label{app:hyperpar}}

\noindent\textbf{WFST model}\quad The Witten--Bell smoothing parameter for the language model is set to 10, and the relative entropy pruning threshold is $10^{-5}$ for the trigram model and $2 \cdot 10^{-5}$ for higher-order models. Unsupervised training is performed in batches of size 10 and the language model order is increased every 100 batches. While training with the bigram model, we disallow insertions and freeze all the deletion probabilities at $e^{-100}$. The EM stepsize decay rate is $\beta = 0.9$. The emission arc pruning threshold is gradually decreased from 5 to 4.5 (in the negative log probability space). We perform multiple random restarts for each experiment, initializing the emission distribution to uniform plus random noise.
\\[0.1in]
\noindent\textbf{UNMT baseline}\quad Our unsupervised neural baseline uses a single-layer LSTM with hidden state size 512 for both the encoder and the decoder. The embedding dimension is set to 128. For the denoising autoencoding loss, we adopt the default noise model and hyperparameters as described by~\citet{lample2018unsupervised}. The autoencoding loss is annealed over the first 3 epochs.

We tune the maximum training sequence length (controlling how much training data is used) and the maximum allowed decoding length by optimizing the validation set CER. In our case, the maximum output length is important because the evaluation metric penalizes the discrepancy in length between the prediction and the reference; we observe the best results when setting it to 40 characters for Arabic and 180 for Russian. At training time, we filter out sequences longer than 100 characters for either language, which constitute 1\% of the available Arabic training data (both the Arabic-only LM training set and the Latin-only training set combined) but almost 70\% of the Russian data. Surprisingly, the Russian model trained on the remaining 30\% achieves better results than the one trained on the full data; we hypothesize that the improvement comes from having a more balanced training set, since the full data is heavily skewed towards the Cyrillic side (LM training set) otherwise (see~\Tref{tab:splits}).

\end{document}